\documentclass{article}

\usepackage{arxiv}
\usepackage{amsmath}
\usepackage[utf8]{inputenc}
\usepackage[T1]{fontenc}
\usepackage{url}
\usepackage{booktabs}
\usepackage{amsfonts}
\usepackage{nicefrac}
\usepackage{microtype}
\usepackage{lipsum}
\usepackage{graphicx}
\usepackage{enumitem}
\usepackage{wrapfig}
\usepackage{multirow}
\usepackage{wasysym}
\usepackage{makecell}
\usepackage{pifont}
\usepackage{colortbl}
\usepackage{caption}
\usepackage[hidelinks]{hyperref}
\usepackage{tabularx}
\usepackage{tcolorbox}
\usepackage{orcidlink}
\usepackage{xcolor}
\usepackage{soul}
\usepackage{array}
\usepackage{calc}
\usepackage{float}
\usepackage{tikz}
\usepackage{natbib}
\usetikzlibrary{shapes.misc}
\usepackage{afterpage}
\graphicspath{ {./images/} }
\definecolor{colorVU}{RGB}{220, 248, 220}
\definecolor{colorDD}{RGB}{255, 225, 195}
\definecolor{colorML}{RGB}{235, 248, 255}
\definecolor{colorQA}{RGB}{250, 235, 250}
\definecolor{colorMM}{RGB}{255, 255, 220}
\definecolor{colorAvg}{RGB}{255, 228, 230}

\begin{document}

\begin{center}
\rule{\textwidth}{2.8pt}  

\vspace{1.1em}

{\Large\bfseries \textit{Med-OPD}: Improving Medical Vision-Language Models via Evidence-Aware On-Policy Distillation}
\vspace{1.1em}
\rule{\textwidth}{1.0pt}  

\vspace{1.2em}  

{\normalsize  
\textbf{Yunhang Qian}$^{1}$ \quad
\textbf {Jiaquan Yu}$^{2}$ \quad
\textbf {Jiawei Liu}$^{2}$ \quad
\textbf {Meng Wang}$^{1}$ \quad
\textbf{Hongwei Bran Li}$^{1}$ \quad
\textbf{Xiaobin Hu}$^{1}$ \quad 
}

\vspace{0.8em}

{\small
$^{1}$National University of Singapore \quad
$^{2}$University of Science and Technology of China \\

}

\end{center}

\vspace{1.5em}
\vspace{1.5em}
\begin{abstract}
Medical Vision-Language Models (Med-VLMs) require reliable reasoning from fine-grained visual evidence, yet existing models can produce plausible clinical answers by relying on language priors or medical templates rather than truly attending to diagnosis-critical regions. On-Policy Distillation (OPD) offers dense token-level supervision on student-generated trajectories and provides a privacy-compatible means of capability transfer without requiring the redistribution of raw patient data. However, standard OPD uniformly distills all tokens, causing sparse evidence-dependent tokens to be diluted by abundant clinical narrative tokens. Inspired by the success of OPD in the large language model community, we propose \textbf{Med-OPD}, to our knowledge the first unified post-training framework that integrates on-policy distillation with medical evidence-aware supervision for Med-VLMs. We introduce \textbf{Medical Evidence Advantage} (MEA), a teacher-grounded counterfactual signal that uses an answer-aware hint to focus teacher scoring on evidence supporting the target diagnosis, and measures each token's dependence on medical visual evidence by comparing teacher likelihoods under the original and evidence-degraded imaging modalities. Based on MEA, Med-OPD redistributes the distillation signal at both the token and trajectory levels, emphasizing diagnosis-critical tokens and evidence-reliant rollouts. Experiments on OmniMedVQA subsets show that Med-OPD consistently outperforms SFT and standard OPD across CT, MRI, Disease Diagnosis, and Lesion Grading. These results demonstrate that evidence-aware distillation can better strengthen medical VLMs' reliance on key visual evidence and improve reliable multimodal medical reasoning. The source code and data is publicly available at: \href{https://github.com/yunhang8658/MedOPD.git}{https://github.com/yunhang8658/MedOPD.git}
\end{abstract}

\section{Introduction}

Medical Vision-Language Models (Med-VLMs) have recently shown great potential in medical visual question answering, report understanding, and computer-aided diagnosis~\cite{li2023llava,chen2024towards,hu2024omnimedvqa,qian2026medmaslab,hu2026landscape,gong2026med,yu2026emambair}. However, medical imaging modality understanding is not merely about recognizing salient semantics in an image; it requires models to make reliable judgments from localized and subtle visual evidence, such as faint infiltrates in chest radiographs or abnormal cellular morphology in pathology slides. Although such evidence occupies only a small portion of the entire image, it often determines the final diagnostic conclusion. Existing medical multimodal models can still rely on question priors, disease co-occurrence patterns, or fixed medical expression templates, producing plausible answers without sufficiently attending to lesion regions~\cite{nguyen2025localizing,liu2024survey,favero2024multi,leng2024mitigating,ghosh2025visual}. This raises a central question: \textbf{How can we make medical VLMs truly rely on key medical visual evidence during reasoning, rather than merely generating fluent medical language?}

\begin{figure}[t]
    \centering
    \includegraphics[width=\linewidth]{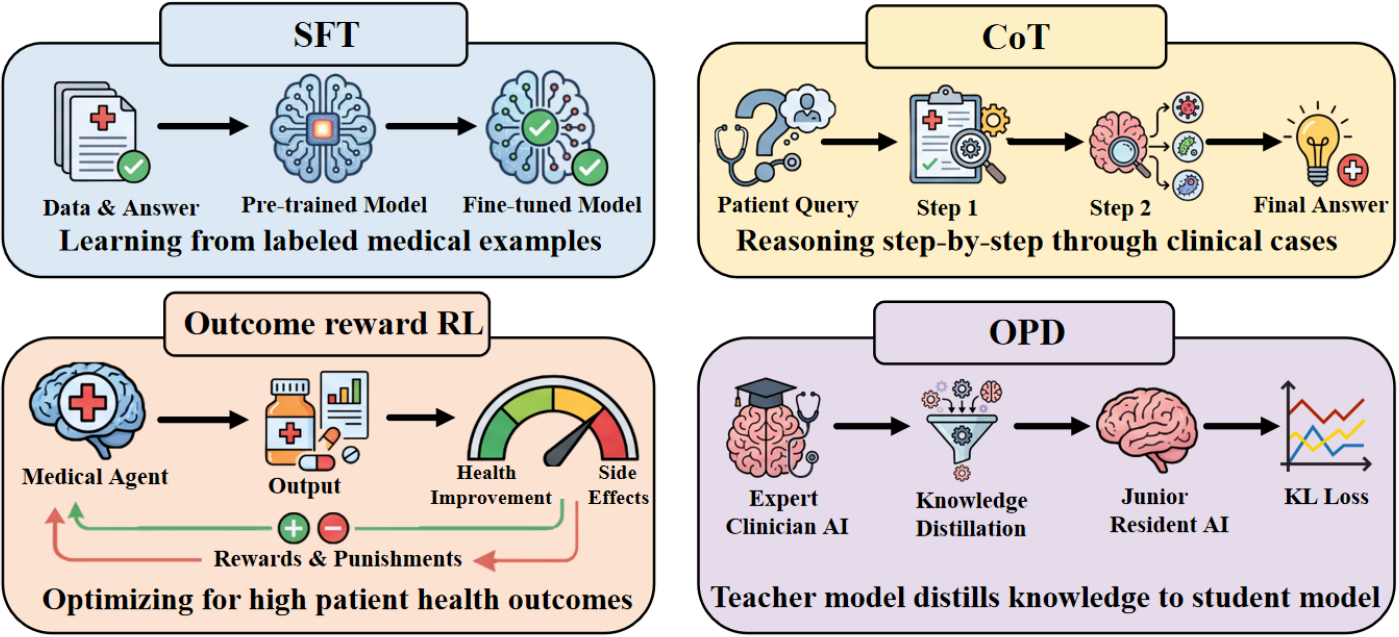}
    \caption{\textbf{Common paradigms for improving medical VLM reasoning.}
    SFT learns from labeled examples, CoT elicits intermediate reasoning steps, outcome-reward RL optimizes scalar feedback, and OPD transfers teacher behavior to a student policy through distillation.}
    \label{fig:paradigms}
\end{figure}

As summarized in Fig.~\ref{fig:paradigms}, existing studies can be roughly grouped into three paradigms. \textbf{(a) Supervised Fine-tuning}, which provides direct supervision through human answers or reasoning trajectories. However, high-quality medical annotations are expensive and difficult to scale, and the mismatch between training and inference distributions in SFT introduces exposure bias, causing early visual misjudgments or language errors to accumulate in subsequent reasoning~\cite{agarwal2024policy,xie2024v}. \textbf{(b) Chain-of-thought and Reinforcement Learning }, where methods such as MedVLM-R1, Med-R1, and MMedExpert-R1 improve medical reasoning and generalization through CoT, GRPO, or clinical guideline rewards~\cite{pan2025medvlm,lai2025medr1,ding2026mmedexpert,shao2024deepseekmath,guo2025deepseek,wei2022chain,li2026joint,guven2026uav,wang2507perception}. Nevertheless, these methods suffer from sparse rewards and training instability, and they mainly optimize final answers, making it difficult to ensure that the reasoning path follows clinical logic and medical evidence. \textbf{(c) On-Policy Distillation}, which combines on-policy sampling with dense token-level distillation, enabling the student to learn the teacher's token-wise distribution on its own generated trajectories~\cite{agarwal2024policy,gu2024minillm,liu2026visual,yu2026dopd,yuan2026vision}. This is attractive for privacy-sensitive medical settings, since capability transfer can rely on teacher distributions rather than redistributing sensitive patient data. However, standard OPD uniformly applies KL distillation to all tokens, while many tokens in medical responses are merely reasoning connectives, diagnostic phrasing, or linguistic scaffolding, and only a few truly depend on visual evidence. This raises the question: \textbf{Can uniform token-level distillation in standard OPD really strengthen the model's visual reliance?}

\begin{figure}[t]
    \centering
    \begin{minipage}[t]{0.48\linewidth}
        \centering
        \includegraphics[width=\linewidth]{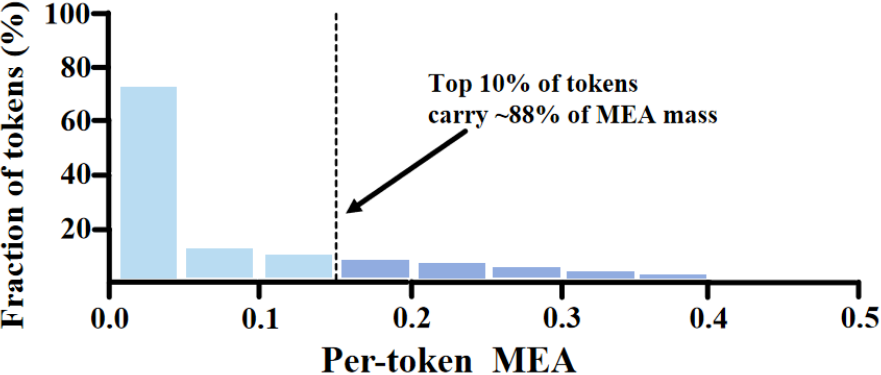}
    \end{minipage}\hfill
    \begin{minipage}[t]{0.48\linewidth}
        \centering
        \includegraphics[width=\linewidth]{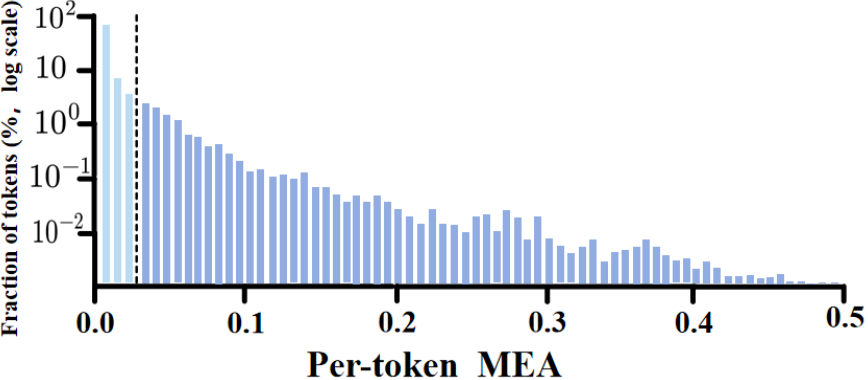}
    \end{minipage}
    \caption{\textbf{MEA is long-tailed.}
    Left: the top $10\%$ high-MEA tokens carry most of the total MEA mass. Right: most tokens have near-zero MEA, while only a small subset strongly depends on medical visual evidence.}
    \label{fig:mea_heavytail}
\end{figure}

\begin{figure}[t]
    \centering
    \includegraphics[width=0.92\linewidth]{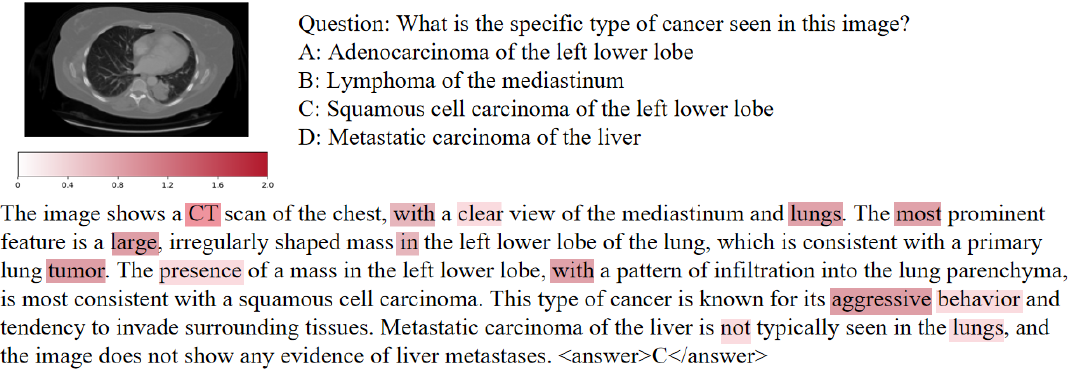}
    \caption{\textbf{High-MEA tokens align with fine-grained medical evidence.}
High-MEA tokens concentrate on clinically relevant details and diagnosis-critical findings, while low-MEA tokens mainly correspond to general narrative.}
    \label{fig:mea_visualization}
\end{figure}

To answer this question, we conduct a token-level analysis on student rollouts from medical VLMs and introduce \textbf{Medical Evidence Advantage} (MEA), a token-level measure of how much a prediction depends on fine-grained medical visual evidence. Specifically, MEA compares the model's confidence in a token before and after the key medical evidence is degraded. As shown in Fig.~\ref{fig:mea_heavytail}, We find that MEA exhibits a pronounced long-tailed distribution. Most tokens have MEA values close to zero, mainly corresponding to linguistic scaffolding in clinical narratives. In Fig.~\ref{fig:mea_visualization}, we find that only a small subset of tokens has high MEA, often corresponding to anatomical locations, lesion morphology, abnormal findings, severity descriptions, and final diagnostic decisions. 

We propose \textbf{Medical Evidence-aware On-Policy Distillation (Med-OPD)}. Unlike standard OPD, which distills all tokens uniformly, Med-OPD uses an answer-aware hint during teacher scoring to direct the teacher toward visual evidence supporting the target diagnosis, and uses MEA to identify evidence-dependent tokens in student rollouts. It then redistributes the distillation signal at both the trajectory and token levels. At the trajectory level, we increase the weight of samples with higher trajectory-averaged MEA, encouraging optimization to focus on samples that genuinely require reading fine-grained medical image details. At the token level, we separately compute KL distillation for high-MEA and low-MEA token groups, preventing key visual tokens such as lesions, anatomical locations, and diagnostic decisions from being overwhelmed by verbose clinical narratives. On the OmniMedVQA benchmark, Med-OPD achieves substantial improvements in both cross-modality and cross-task settings, raising the average accuracy by 5.24 percentage points over SFT and 2.83 percentage points over standard OPD, demonstrating that evidence-aware distillation more effectively strengthens medical visual reasoning.

Our main contributions are summarized as follows:

\begin{itemize}
    \item We identify an evidence utilization gap in medical VLMs and show that standard OPD suffers from visual supervision dilution in the medical setting: medical evidence is highly concentrated on a small number of diagnosis-critical tokens, while uniform token-level KL weakens their training signal.
    
    \item We introduce an answer-aware teacher hint that directs teacher scoring toward visual findings supporting the target diagnosis, and define \textbf{Medical Evidence Advantage} (MEA) to quantify each token's dependence on such evidence. We empirically find that MEA follows a pronounced long-tailed distribution in medical student rollouts, where the top $10\%$ high-MEA tokens carry the vast majority of the total MEA mass.                      
    
    \item We propose \textbf{Med-OPD}, a medical evidence-aware on-policy distillation framework that combines answer-aware teacher scoring with trajectory- and token-level signal redistribution, thereby strengthening the model's reliance on key medical visual evidence and validating its effectiveness on medical multimodal question answering tasks.
\end{itemize}

\section{Related Work}

\subsection{Medical Vision-Language Models and Medical Reasoning}

Med-VLMs have rapidly advanced with the development of domain-specific multimodal instruction tuning and large-scale medical benchmarks. Representative models such as LLaVA-Med and HuatuoGPT-Vision adapt general VLMs to biomedical image-text understanding through medical instruction data and domain-specific alignment~\cite{li2023llava,chen2024towards}. Meanwhile, benchmarks such as OmniMedVQA provide large-scale evaluations across diverse medical modalities and anatomical regions, revealing that current Med-VLMs still struggle with robust medical visual reasoning~\cite{hu2024omnimedvqa}. Recent works further introduce reasoning-oriented post-training methods, including MedVLM-R1, Med-R1, and MMedExpert-R1, to improve medical reasoning and generalization through reinforcement learning or clinical guideline alignment~\cite{pan2025medvlm,lai2025medr1,ding2026mmedexpert}. However, these methods mainly optimize answer correctness or reasoning format, while the model's reliance on fine-grained medical visual evidence remains insufficiently explored.

\subsection{Grounding and Hallucination in Medical VLMs}

Hallucination is a critical failure mode for VLMs, where models generate fluent but visually unsupported content~\cite{liu2024survey,li2023evaluating,li2023blip}. This issue is especially risky in medicine, since plausible clinical language may conceal the absence of true visual grounding. Recent studies show that medical multimodal models often rely on textual shortcuts or dataset biases rather than localizing the relevant pathological regions. HEAL-MedVQA framework highlights this problem by evaluating whether models answer based on grounded lesion evidence~\cite{nguyen2025localizing,liu2024improved,liu2023visual}. These findings suggest that improving medical VLMs requires not only better final accuracy, but also stronger dependence on the visual evidence that supports diagnosis-critical tokens.

\subsection{On-Policy Distillation and Evidence-Aware Supervision}

OPD offers a promising alternative to supervised fine-tuning and reward-based reinforcement learning by matching teacher distributions on student-generated rollouts, which provides dense and stable token-level supervision~\cite{agarwal2024policy,gu2024minillm,wang2026openclaw}. This distribution-level transfer is particularly appealing in medical AI, where raw patient data and expert reasoning traces are privacy-sensitive. However, standard OPD applies uniform KL distillation to all tokens, even though only a few tokens in medical responses are grounded in visual evidence. Visual-Advantage OPD identifies vision-critical tokens in general VLMs through teacher-scored visual advantage~\cite{liu2026visual}. In medical settings, we use an answer-aware hint to guide teacher scoring toward findings that support the target diagnosis, introduce Medical Evidence Advantage, and propose Med-OPD to prioritize diagnosis-critical tokens.

\section{Method}

\subsection{On-Policy Distillation for Medical VLMs}

For two token distributions $p$ and $q$ over the vocabulary $\mathcal{V}$, the Kullback--Leibler (KL) divergence~\cite{ko2024distillm} is defined as
\begin{equation}
    \mathrm{KL}(p\|q)
    =
    \sum_{w\in\mathcal{V}} p(w)
    \log \frac{p(w)}{q(w)}.
    \label{eq:kl}
\end{equation}
Given a training triple $(v,q,y)\sim\mathcal{S}$, where $v$ is a medical image, $q$ is a question, and $y=(y_1,\ldots,y_{|y|})$ is the target answer, supervised fine-tuning (SFT) trains the student by teacher forcing:
\begin{equation}
    \mathcal{L}_{\mathrm{SFT}}
    =
    -\mathbb{E}_{(v,q,y)\sim\mathcal{S}}
    \left[
    \frac{1}{|y|}
    \sum_{t=1}^{|y|}
    \log \pi_{\theta_S}(y_t \mid v,q,y_{<t})
    \right].
    \label{eq:sft}
\end{equation}
During SFT, the student always conditions on the reference prefix $y_{<t}$. At inference, however, it must condition on its own prefix $\hat{y}_{<t}$, creating a state-distribution mismatch. In medical reasoning, an early visual misinterpretation can therefore shift the subsequent narrative away from the training distribution and compound across tokens~\cite{ross2011reduction,xu2025speculative,li2025self}.

OPD addresses this discrepancy by querying the teacher on prefixes generated by the student itself. We use $q'$ to denote an answer-aware teacher prompt constructed by augmenting $q$ with the correct answer. The student policy $\pi_{\theta_S}(\cdot \mid v,q)$ samples an on-policy rollout $\hat{y}=(\hat{y}_1,\ldots,\hat{y}_{|\hat{y}|})$, while the teacher policy $\pi_{\theta_T}(\cdot \mid v,q',\hat{y}_{<t})$ provides token-level supervision on the same student-generated prefixes. Standard OPD minimizes the expected reverse-KL divergence:
\begin{equation}
    \mathcal{L}_{\mathrm{OPD}}
    =
    \mathbb{E}_{(v,q)\sim\mathcal{S},\, \hat{y}\sim\pi_{\theta_S}(\cdot\mid v,q)}
    \left[
    \frac{1}{|\hat{y}|}
    \sum_{t=1}^{|\hat{y}|}
    \mathrm{KL}
    \left(
    \pi_{\theta_S}(\cdot \mid v,q,\hat{y}_{<t})
    \,\|\, 
    \pi_{\theta_T}(\cdot \mid v,q',\hat{y}_{<t})
    \right)
    \right],
    \label{eq:opd}
\end{equation}
Compared with SFT, OPD aligns supervision with the states visited by the student during generation. It also provides denser and more stable token-level training signals than reinforcement learning based only on sparse final-answer rewards.

\begin{figure*}[t]
    \centering
    \includegraphics[width=0.98\textwidth]{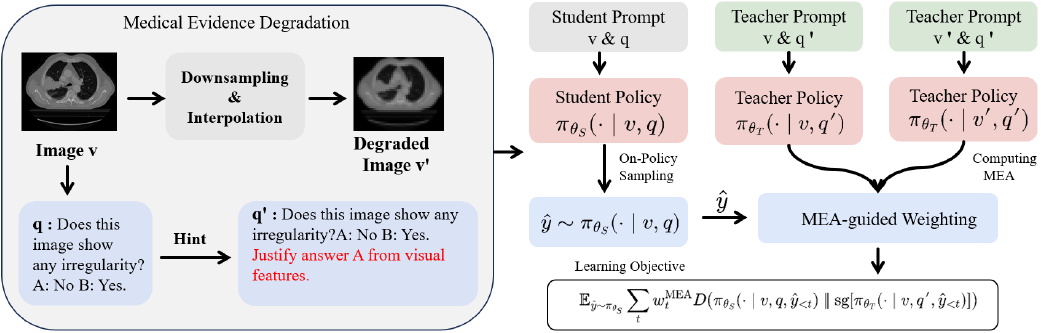}
    \caption{\textbf{Overview of Med-OPD.}
Med-OPD first constructs an evidence-degraded view of the medical image and an answer-aware teacher prompt. The student generates on-policy rollouts from the original image-question pair, while the teacher evaluates these rollouts under both full-evidence and degraded-evidence conditions to estimate MEA. The resulting MEA signal is used to reallocate distillation strength across rollouts and tokens, encouraging the model to focus on evidence-reliant samples and diagnosis-critical visual tokens.}
    \label{fig:arch}
\end{figure*}

\subsection{Medical Evidence Advantage}

To identify evidence-dependent tokens in student rollouts, we introduce \textbf{Medical Evidence Advantage}. The key idea is a teacher-grounded counterfactual comparison: if a token becomes much more likely when fine-grained medical evidence is available, and less likely when such evidence is weakened, then this token is more likely to carry diagnosis-relevant visual information.

As shown in Fig.~\ref{fig:arch}, given the original medical image $v$, we construct an evidence-degraded image $v'$ by first downsampling $v$ to $10\%$ of its original spatial resolution and then resizing it back to the original size using bilinear interpolation. This degradation weakens local fine-grained medical evidence, such as lesion texture, subtle boundaries, tiny abnormalities, or local density changes, while largely preserving global anatomy and modality information. Meanwhile, the original question $q$ is augmented into an answer-aware teacher prompt $q'$ by incorporating the target answer instruction. For the $t$-th token in a student-generated rollout $\hat{y}$, MEA is defined as:
\begin{equation}
    a_t
    =
    \max\left(
    \log \pi_{\theta_T}(\hat{y}_t \mid v,q',\hat{y}_{<t})
    -
    \log \pi_{\theta_T}(\hat{y}_t \mid v',q',\hat{y}_{<t}),
    0
    \right),
    \label{eq:mea}
\end{equation}
A larger $a_t$ indicates that the teacher becomes more confident in predicting this token when full medical visual evidence is available, suggesting stronger dependence on medical evidence. In contrast, a near-zero $a_t$ usually corresponds to language priors, reasoning templates, or generic medical expressions.

\subsection{MEA-guided Token-level Grouping}

Standard OPD averages the distillation divergence over all generated tokens. In medical answers, however, only a small portion of tokens directly expresses visual evidence, such as lesion type, anatomical site, abnormal finding, or diagnostic decision, while many other tokens mainly connect the clinical narrative. A uniform token average can therefore make the supervision from evidence-critical tokens vanish inside long responses. Med-OPD addresses this issue by separating tokens according to their MEA scores and computing the distillation loss within each group.

For a student rollout $\hat{y}$ of length $T=|\hat{y}|$, we use the MEA scores $\{a_t\}_{t=1}^{T}$ computed in Eq.~\ref{eq:mea} to identify evidence-critical tokens. Let $\sigma$ be a permutation of token indices that sorts these scores in descending order, i.e., $a_{\sigma(1)}\geq a_{\sigma(2)}\geq\cdots\geq a_{\sigma(T)}$. We select the first $m=\lceil \rho T\rceil$ indices as the high-evidence group and assign the remaining indices to the low-evidence group:
\begin{equation}
    \mathcal{H}(\hat{y})
    =
    \{\sigma(1),\ldots,\sigma(m)\},
    \qquad
    \mathcal{L}(\hat{y})
    =
    \{\sigma(m+1),\ldots,\sigma(T)\}.
    \label{eq:token_groups}
\end{equation}
Here $\rho\in(0,1)$ controls the proportion of tokens assigned to the high-evidence group.

After the token groups are determined, we compute the per-token distillation divergence
\begin{equation}
    KL_t
    =
    KL
    \left(
    \pi_{\theta_S}(\cdot \mid v,q,\hat{y}_{<t})
    \,\|\, 
    \pi_{\theta_T}(\cdot \mid v,q',\hat{y}_{<t})
    \right).
    \label{eq:token_div}
\end{equation}
Instead of averaging $KL_t$ over all tokens in the rollout at once, we average the two groups separately and combine them with a balancing coefficient $\lambda$:
\begin{equation}
    \mathcal{L}_{\mathrm{group}}(\hat{y})
    =
    \lambda
    \frac{1}{|\mathcal{H}(\hat{y})|}
    \sum_{t\in\mathcal{H}(\hat{y})} KL_t
    +
    (1-\lambda)
    \frac{1}{|\mathcal{L}(\hat{y})|}
    \sum_{t\in\mathcal{L}(\hat{y})} KL_t .
    \label{eq:group_loss}
\end{equation}
This grouped formulation prevents the high-MEA group from being dominated by the much larger number of low-MEA tokens. As a result, diagnosis-critical visual tokens receive a stable share of the distillation signal even when the response contains many generic clinical expressions.

\subsection{MEA-guided Trajectory-level Weighting}

For the same medical image-question pair, we draw $K$ on-policy rollouts $\{\hat{y}^{(k)}\}_{k=1}^{K}$, which may follow different reasoning paths and are not equally useful for evidence-oriented learning. A response that consistently refers to image-grounded findings should contribute more to training than one dominated by generic clinical wording. Applying Eq.~\ref{eq:mea} to the $k$-th rollout gives token scores $\{a_t^{(k)}\}_{t=1}^{T^{(k)}}$, whose average summarizes its evidence dependence:
\begin{equation}
    \bar{a}^{(k)}
    =
    \frac{1}{T^{(k)}}
    \sum_{t=1}^{T^{(k)}} a_t^{(k)},
    \qquad
    \hat{z}^{(k)}
    =
    \frac{\bar{a}^{(k)}-\mu}{\sigma+\epsilon},
    \label{eq:traj_mea}
\end{equation}
where $T^{(k)}=|\hat{y}^{(k)}|$. The mean $\mu$ and standard deviation $\sigma$ are computed over the $K$ rollout-level MEA scores from the same input, and $\epsilon$ is a small constant. This normalization makes the score reflect how evidence-reliant a rollout is relative to its alternatives for the same case. We obtain the rollout weight by applying a temperature-controlled softmax:
\begin{equation}
    w^{(k)}
    =
    \frac{\exp(\hat{z}^{(k)} / \tau)}
    {\sum_{j=1}^{K}\exp(\hat{z}^{(j)} / \tau)},
    \qquad
    \sum_{k=1}^{K} w^{(k)} = 1 .
    \label{eq:traj_weight}
\end{equation}
In this way, trajectory-level weighting is performed within each candidate set, so the model places more optimization weight on the rollout that better reflects the medical evidence in the image.

\subsection{Final Objective}

Combining token-level grouping and trajectory-level weighting gives the per-sample Med-OPD loss:
\begin{equation}
    \mathcal{L}_{\mathrm{Med\text{-}OPD}}(v,q)
    =
    \sum_{k=1}^{K}
    w^{(k)}
    \mathcal{L}_{\mathrm{group}}(\hat{y}^{(k)}).
    \label{eq:medevi_opd}
\end{equation}
The full training objective is $\mathbb{E}_{(v,q)\sim\mathcal{S}}\left[\mathcal{L}_{\mathrm{Med\text{-}OPD}}(v,q)\right]$, where the $K$ rollouts are sampled on-policy from the student for each training sample.
Importantly, the evidence-degraded image $v'$ is used only as a counterfactual reference for computing MEA. The final teacher target in $KL_t$ is always conditioned on the original image $v$ and the answer-aware prompt $q'$, ensuring that Med-OPD distills from the teacher's full-evidence prediction while using MEA to decide where the distillation signal should be concentrated.

\begin{table}[t]
\centering
\caption{\textbf{Results on OmniMedVQA subsets.} We evaluate on two modality subsets (CT and MRI) and two task-type subsets (Disease Diagnosis and Lesion Grading). The best and second-best results in each column are bolded and underlined, respectively.}
\label{tab:main_results}
\begin{tabular}{lccccc}
\toprule
\textbf{Model} & \textbf{CT} & \textbf{MRI} & \textbf{Disease Diagnosis} & \textbf{Lesion Grading} & \textbf{Avg.} \\
\midrule
\multicolumn{6}{c}{\textit{Zero-shot General VLMs}} \\
Qwen3-VL-2B  & 0.5867 & 0.6700 & 0.5933 & 0.7000 & 0.6375 \\
Qwen3-VL-4B  & \underline{0.7033} & 0.7200 & 0.7433 & \textbf{0.8000} & \textbf{0.7417} \\
Qwen3-VL-8B  & \textbf{0.7100} & 0.7367 & 0.6600 & 0.6900 & 0.6992 \\
Qwen3-VL-32B & 0.6867 & 0.7533 & 0.7000 & 0.7067 & 0.7117 \\
\midrule
\multicolumn{6}{c}{\textit{Zero-shot Medical VLMs}} \\
HealthGPT-Pro-4B & 0.6700 & \textbf{0.8333} & \textbf{0.7967} & 0.6567 & \underline{0.7392} \\
HealthGPT-Pro-8B & 0.6900 & \underline{0.8000} & \underline{0.7867} & 0.6367 & 0.7284 \\
\midrule
\multicolumn{6}{c}{\textit{Fine-tuned VLMs}} \\
Qwen3-VL-2B (SFT)  & 0.6067 & 0.6800 & 0.6000 & 0.7267 & 0.6534 \\
Qwen3-VL-2B (OPD)  & 0.6167 & 0.7167 & 0.6267 & 0.7500 & 0.6775 \\
Qwen3-VL-2B (Ours) & 0.6300 & 0.7667 & 0.6533 & \underline{0.7733} & 0.7058 \\
\bottomrule
\end{tabular}
\end{table}

\section{Experiments }

\subsection{Experimental Setup}

\textbf{Dataset.} We conduct experiments on OmniMedVQA~\cite{hu2024omnimedvqa}, a large-scale medical visual question answering benchmark covering diverse imaging modalities and clinical question types. Following our goal of evaluating whether evidence-aware OPD can transfer to medical scenarios, we build two groups of controlled subsets. For cross-modality evaluation, we select the CT and MRI categories and randomly sample $1{,}000$ training examples and $300$ test examples from each modality. For cross-task evaluation, we select Disease Diagnosis and Lesion Grading and similarly sample $1{,}000$ training examples and $300$ test examples for each task type. These subsets focus on settings where subtle visual evidence, such as lesion morphology, anatomical location, and abnormal tissue appearance, is important for reliable reasoning.

\textbf{Training.} Training is conducted on an AMD server with four MI308 GPUs, each equipped with $192$GB memory. We implement training with PyTorch, vLLM, and FlashAttention-2. Unless otherwise specified, we use a learning rate of $2\times10^{-6}$, a batch size of $256$, and bfloat16 mixed precision. Each training run lasts for up to $10$ epochs, with a maximum of $100$ or $200$ training steps depending on the experimental setting, and evaluation is performed every $10$ training steps. For OPD-based methods, we use Student Top-K distillation with $k_{\mathrm{top}}=16$~\cite{li2026rethinking}. For Med-OPD, the student samples $K=8$ candidate answers for each training example, which are used to compute MEA-guided trajectory-level weights and token-level grouped distillation losses.

\textbf{Baseline Methods \& Evaluation Metric.} We compare Med-OPD with three groups of baselines. First, we evaluate general-purpose zero-shot VLMs, including Qwen3-VL-2B-Instruct, Qwen3-VL-4B-Instruct, Qwen3-VL-8B-Instruct, and Qwen3-VL-32B-Instruct. Second, we include zero-shot medical VLMs, including HealthGPT-Pro-4B and HealthGPT-Pro-8B. Third, we compare against fine-tuned Qwen3-VL-2B-Instruct variants trained with supervised fine-tuning and standard OPD. All methods are evaluated using accuracy, where a prediction is considered correct if the selected answer option matches the ground-truth answer.

\subsection{Results and Analysis}

We use Qwen3-VL-2B-Instruct as the student backbone for all fine-tuned methods. For both standard OPD and Med-OPD, we use Qwen3-VL-4B-Instruct as the teacher model. Table~\ref{tab:main_results} shows that standard OPD already improves the 2B student over SFT by 2.41 percentage points on average, supporting the value of learning from student-generated states. Med-OPD further improves the average accuracy from 0.6775 to 0.7058 over standard OPD (2.83 percentage points), and from 0.6534 to 0.7058 over SFT (5.24 percentage points), with gains on all four subsets.

The largest gain over standard OPD occurs on MRI (0.7167 to 0.7667), where subtle anatomical and lesion-level cues are particularly important. Improvements on Disease Diagnosis and Lesion Grading further indicate that MEA-guided grouping benefits both diagnostic decisions and severity-related reasoning by preventing diagnosis-critical tokens from being diluted by generic clinical narratives.

\begin{figure}[t]
    \centering
    \includegraphics[width=\linewidth]{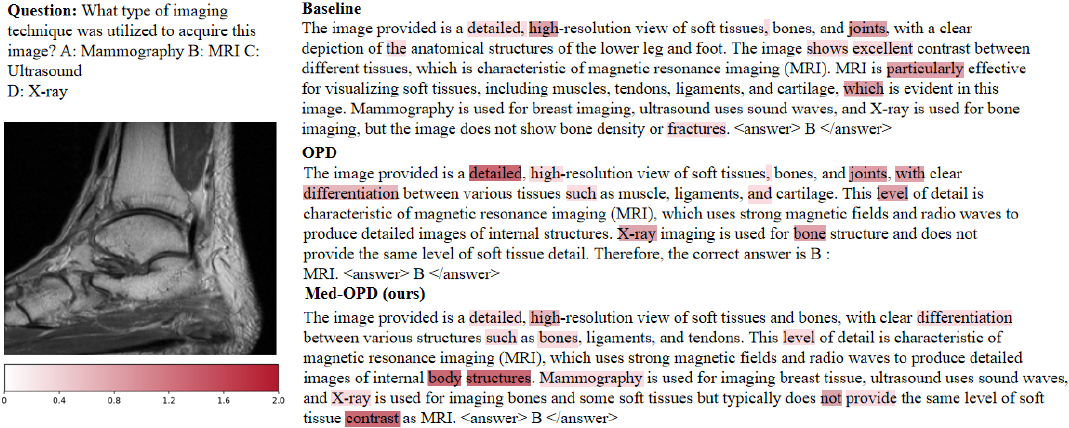}
    \caption{\textbf{Qualitative comparison of evidence-grounded reasoning.}
    Token-level highlights on a representative MRI modality question. Med-OPD more strongly emphasizes image-grounded descriptors, including internal body structures and soft-tissue contrast, than the baseline and standard OPD.}
    \label{fig:qualitative_compare}
\end{figure}

Figure~\ref{fig:qualitative_compare} illustrates this behavior on a representative MRI modality question. Compared with the baseline and standard OPD, Med-OPD emphasizes image-grounded descriptions such as internal body structures and soft-tissue contrast while preserving the correct modality prediction.

Finally, larger zero-shot general or medical VLMs remain strong on some subsets, but parameter scale or medical pretraining alone does not ensure evidence reliance. The gains obtained by Med-OPD with the 2B student therefore suggest that evidence-aware post-training is complementary to model scaling: it changes how supervision is allocated toward tokens grounded in medical visual evidence.

\section{Conclusion}

In this work, we investigate how on-policy distillation can be adapted to medical vision-language models, where reliable reasoning often depends on sparse, localized, and diagnosis-critical visual evidence. OPD is particularly appealing in medical settings because it transfers teacher behavior through output distributions on student-generated rollouts, reducing the need to redistribute privacy-sensitive patient data or expert reasoning traces. However, we show that standard OPD can dilute medical evidence by distributing token-level supervision uniformly across long clinical narratives. To address this issue, we propose Med-OPD, which uses an answer-aware teacher hint to assess visual findings supporting the target diagnosis and introduces Medical Evidence Advantage to identify evidence-dependent tokens. Med-OPD then reallocates distillation signals at both the trajectory and token levels, emphasizing evidence-reliant rollouts and diagnosis-critical tokens. Experiments on controlled OmniMedVQA subsets demonstrate consistent improvements over SFT and standard OPD across CT, MRI, Disease Diagnosis, and Lesion Grading, improving average accuracy by 5.24 and 2.83 percentage points, respectively. These results suggest that effective medical VLM post-training should not only provide dense supervision, but also explicitly account for where clinically meaningful visual evidence is concentrated.

\bibliography{references}

@String(JMLR  = {J. Mach. Learn. Res.})

@String(JMLR  = {JMLR})

@article{nguyen2025localizing,
  title={Localizing Before Answering: A Hallucination Evaluation Benchmark for Grounded Medical Multimodal LLMs},
  author={Nguyen, Dung and Ho, Minh Khoi and Ta, Huy and Nguyen, Thanh Tam and Chen, Qi and Rav, Kumar and Dang, Quy Duong and Ramchandre, Satwik and Phung, Son Lam and Liao, Zhibin and others},
  journal={arXiv preprint arXiv:2505.00744},
  year={2025}
}

@article{lai2025medr1,
  title={Med-r1: Reinforcement learning for generalizable medical reasoning in vision-language models},
  author={Lai, Yuxiang and Zhong, Jike and Li, Ming and Zhao, Shitian and Li, Yuheng and Psounis, Konstantinos and Yang, Xiaofeng},
  journal={IEEE transactions on medical imaging},
  year={2026},
  publisher={IEEE}
}

@inproceedings{pan2025medvlm,
  title={Medvlm-r1: Incentivizing medical reasoning capability of vision-language models (vlms) via reinforcement learning},
  author={Pan, Jiazhen and Liu, Che and Wu, Junde and Liu, Fenglin and Zhu, Jiayuan and Li, Hongwei Bran and Chen, Chen and Ouyang, Cheng and Rueckert, Daniel},
  booktitle={International Conference on Medical Image Computing and Computer-Assisted Intervention},
  pages={337--347},
  year={2025},
  organization={Springer}
}

@article{ding2026mmedexpert,
  title={MMedExpert-R1: Strengthening Multimodal Medical Reasoning via Domain-Specific Adaptation and Clinical Guideline Reinforcement},
  author={Ding, Meidan and Zhang, Jipeng and Wang, Wenxuan and Zhong, Haiqin and Luo, Xiaoling and Chen, Wenting and Shen, Linlin},
  journal={arXiv preprint arXiv:2601.10949},
  year={2026}
}

@article{liu2026visual,
  title={Visual-Advantage On-Policy Distillation for Vision-Language Models},
  author={Liu, Ruiqi and Lv, Xiaolei and Li, Gengsheng and Zhu, Ximo and Wang, Zhiheng and Zhang, Zhengbo and Chen, Junkai and Li, Zhiheng and Li, Bo and Gao, Jun and others},
  journal={arXiv preprint arXiv:2605.21924},
  year={2026}
}

@inproceedings{hu2024omnimedvqa,
  title={Omnimedvqa: A new large-scale comprehensive evaluation benchmark for medical lvlm},
  author={Hu, Yutao and Li, Tianbin and Lu, Quanfeng and Shao, Wenqi and He, Junjun and Qiao, Yu and Luo, Ping},
  booktitle={Proceedings of the IEEE/CVF Conference on Computer Vision and Pattern Recognition},
  pages={22170--22183},
  year={2024}
}

@article{guo2025deepseek,
  title={Deepseek-r1: Incentivizing reasoning capability in llms via reinforcement learning},
  author={Guo, Daya and Yang, Dejian and Zhang, Haowei and Song, Junxiao and Wang, Peiyi and Zhu, Qihao and Xu, Runxin and Zhang, Ruoyu and Ma, Shirong and Bi, Xiao and others},
  journal={arXiv preprint arXiv:2501.12948},
  year={2025}
}

@article{shao2024deepseekmath,
  title={Deepseekmath: Pushing the limits of mathematical reasoning in open language models, 2024},
  author={Shao, Zhihong and Wang, Peiyi and Zhu, Qihao and Xu, Runxin and Song, Junxiao and Bi, Xiao and Zhang, Haowei and Zhang, Mingchuan and Li, YK and Wu, Yang and others},
  journal={URL https://arxiv. org/abs/2402.03300},
  volume={2},
  number={3},
  pages={5},
  year={2024}
}

@inproceedings{agarwal2024policy,
  title={On-policy distillation of language models: Learning from self-generated mistakes},
  author={Agarwal, Rishabh and Vieillard, Nino and Zhou, Yongchao and Stanczyk, Piotr and Ramos Garea, Sabela and Geist, Matthieu and Bachem, Olivier},
  booktitle={International Conference on Learning Representations},
  volume={2024},
  pages={21246--21263},
  year={2024}
}

@inproceedings{gu2024minillm,
  title={Minillm: Knowledge distillation of large language models},
  author={Gu, Yuxian and Dong, Li and Wei, Furu and Huang, Minlie},
  booktitle={International Conference on Learning Representations},
  volume={2024},
  pages={32694--32717},
  year={2024}
}

@article{li2023llava,
  title={Llava-med: Training a large language-and-vision assistant for biomedicine in one day},
  author={Li, Chunyuan and Wong, Cliff and Zhang, Sheng and Usuyama, Naoto and Liu, Haotian and Yang, Jianwei and Naumann, Tristan and Poon, Hoifung and Gao, Jianfeng},
  journal={Advances in Neural Information Processing Systems},
  volume={36},
  pages={28541--28564},
  year={2023}
}

@inproceedings{li2023blip,
  title={Blip-2: Bootstrapping language-image pre-training with frozen image encoders and large language models},
  author={Li, Junnan and Li, Dongxu and Savarese, Silvio and Hoi, Steven},
  booktitle={International conference on machine learning},
  pages={19730--19742},
  year={2023},
  organization={PMLR}
}

@inproceedings{li2023evaluating,
  title={Evaluating object hallucination in large vision-language models},
  author={Li, Yifan and Du, Yifan and Zhou, Kun and Wang, Jinpeng and Zhao, Xin and Wen, Ji-Rong},
  booktitle={Proceedings of the 2023 conference on empirical methods in natural language processing},
  pages={292--305},
  year={2023}
}

@inproceedings{chen2024towards,
  title={Towards injecting medical visual knowledge into multimodal llms at scale},
  author={Chen, Junying and Gui, Chi and Ouyang, Ruyi and Gao, Anningzhe and Chen, Shunian and Chen, Guiming Hardy and Wang, Xidong and Cai, Zhenyang and Ji, Ke and Wan, Xiang and others},
  booktitle={Proceedings of the 2024 conference on empirical methods in natural language processing},
  pages={7346--7370},
  year={2024}
}

@article{liu2024survey,
  title={A survey on hallucination in large vision-language models},
  author={Liu, Hanchao and Xue, Wenyuan and Chen, Yifei and Chen, Dapeng and Zhao, Xiutian and Wang, Ke and Hou, Liping and Li, Rongjun and Peng, Wei},
  journal={arXiv preprint arXiv:2402.00253},
  year={2024}
}

@article{li2026rethinking,
  title={Rethinking on-policy distillation of large language models: Phenomenology, mechanism, and recipe},
  author={Li, Yaxuan and Zuo, Yuxin and He, Bingxiang and Zhang, Jinqian and Xiao, Chaojun and Qian, Cheng and Yu, Tianyu and Gao, Huan-ang and Yang, Wenkai and Liu, Zhiyuan and others},
  journal={arXiv preprint arXiv:2604.13016},
  year={2026}
}

@article{yu2026dopd,
  title={DOPD: Dual On-policy Distillation},
  author={Yu, Xinlei and Li, Gen and Si, Qingyi and Zhang, Guibin and Xu, Yuqi and Wang, Congcong and Dong, Shuai and Tuo, Kaiwen and Zeng, Xiangyu and Feng, Kaituo and others},
  journal={arXiv preprint arXiv:2606.30626},
  year={2026}
}

@article{qian2026medmaslab,
  title={Medmaslab: A unified orchestration framework for benchmarking multimodal medical multi-agent systems},
  author={Qian, Yunhang and Hu, Xiaobin and Yu, Jiaquan and Xin, Siyang and Chen, Xiaokun and Zhang, Jiangning and Jiang, Peng-Tao and Liu, Jiawei and Li, Hongwei Bran},
  journal={arXiv preprint arXiv:2603.09909},
  year={2026}
}

@article{hu2026landscape,
  title={The landscape of medical agents: A survey},
  author={Hu, Xiaobin and Qian, Yunhang and Yu, Jiaquan and Liu, Jingjing and Ji, Xiaozhong and Xu, Chengming and Tang, Peng and Xu, Chengming and Tang, Peng and Liu, Jiawei and others},
  year={2026},
  publisher={TechRxiv}
}

@article{wei2022chain,
  title={Chain-of-thought prompting elicits reasoning in large language models},
  author={Wei, Jason and Wang, Xuezhi and Schuurmans, Dale and Bosma, Maarten and Xia, Fei and Chi, Ed and Le, Quoc V and Zhou, Denny and others},
  journal={Advances in neural information processing systems},
  volume={35},
  pages={24824--24837},
  year={2022}
}

@inproceedings{gong2026med,
  title={Med-cmr: A fine-grained benchmark integrating visual evidence and clinical logic for medical complex multimodal reasoning},
  author={Gong, Haozhen and Ji, Xiaozhong and Liu, Yuansen and Wu, Wenbin and Yan, Xiaoxiao and Liu, Jingjing and Wu, Kai and Pan, Jiazhen and Jian, Bailiang and Zhang, Jiangning and others},
  booktitle={Proceedings of the IEEE/CVF Conference on Computer Vision and Pattern Recognition},
  pages={41224--41234},
  year={2026}
}

@article{yu2026emambair,
  title={EmambaIR: Efficient Visual State Space Model for Event-guided Image Reconstruction},
  author={Yu, Wei and Qian, Yunhang},
  journal={arXiv preprint arXiv:2605.08073},
  year={2026}
}

@article{yuan2026vision,
  title={Vision-opd: Learning to see fine details for multimodal llms via on-policy self-distillation},
  author={Yuan, Qianhao and Lou, Jie and Yu, Xing and Lin, Hongyu and Sun, Le and Han, Xianpei and Lu, Yaojie},
  journal={arXiv preprint arXiv:2605.18740},
  year={2026}
}

@article{ko2024distillm,
  title={Distillm: Towards streamlined distillation for large language models},
  author={Ko, Jongwoo and Kim, Sungnyun and Chen, Tianyi and Yun, Se-Young},
  journal={arXiv preprint arXiv:2402.03898},
  year={2024}
}

@inproceedings{liu2024improved,
  title={Improved baselines with visual instruction tuning},
  author={Liu, Haotian and Li, Chunyuan and Li, Yuheng and Lee, Yong Jae},
  booktitle={Proceedings of the IEEE/CVF conference on computer vision and pattern recognition},
  pages={26296--26306},
  year={2024}
}

@article{liu2023visual,
  title={Visual instruction tuning},
  author={Liu, Haotian and Li, Chunyuan and Wu, Qingyang and Lee, Yong Jae},
  journal={Advances in neural information processing systems},
  volume={36},
  pages={34892--34916},
  year={2023}
}

@article{li2026joint,
  title={Joint Optimization of Reasoning and Dual-Memory for Self-Learning Diagnostic Agent},
  author={Li, Bingxuan and Du, Simo and Guo, Yue},
  journal={arXiv preprint arXiv:2604.07269},
  year={2026}
}

@article{guven2026uav,
  title={UAV-MARL: Multi-Agent Reinforcement Learning for Time-Critical and Dynamic Medical Supply Delivery},
  author={Guven, Islam and Parlak, Mehmet},
  journal={arXiv preprint arXiv:2603.10528},
  year={2026}
}

@inproceedings{ross2011reduction,
  title={A reduction of imitation learning and structured prediction to no-regret online learning},
  author={Ross, St{\'e}phane and Gordon, Geoffrey and Bagnell, Drew},
  booktitle={Proceedings of the fourteenth international conference on artificial intelligence and statistics},
  pages={627--635},
  year={2011},
  organization={JMLR Workshop and Conference Proceedings}
}

@inproceedings{xu2025speculative,
  title={Speculative knowledge distillation: Bridging the teacher-student gap through interleaved sampling},
  author={Xu, Wenda and Han, Rujun and Wang, Zifeng and Le, Long and Madeka, Dhruv and Li, Lei and Wang, William and Agarwal, Rishabh and Lee, Chen-Yu and Pfister, Tomas},
  booktitle={International Conference on Learning Representations},
  volume={2025},
  pages={64616--64646},
  year={2025}
}

@inproceedings{xie2024v,
  title={V-dpo: Mitigating hallucination in large vision language models via vision-guided direct preference optimization},
  author={Xie, Yuxi and Li, Guanzhen and Xu, Xiao and Kan, Min-Yen},
  booktitle={Findings of the Association for Computational Linguistics: EMNLP 2024},
  pages={13258--13273},
  year={2024}
}

@article{wang2026openclaw,
  title={Openclaw-rl: Train any agent simply by talking},
  author={Wang, Yinjie and Chen, Xuyang and Jin, Xiaolong and Wang, Mengdi and Yang, Ling},
  journal={arXiv preprint arXiv:2603.10165},
  year={2026}
}

@article{wang2507perception,
  title={Perception-aware policy optimization for multimodal reasoning (2025)},
  author={Wang, Zhenhailong and Guo, Xuehang and Stoica, Sofia and Xu, Haiyang and Wang, Hongru and Ha, Hyeonjeong and Chen, Xiusi and Chen, Yangyi and Yan, Ming and Huang, Fei and others},
  journal={URL https://arxiv. org/abs/2507.06448}
}

@inproceedings{favero2024multi,
  title={Multi-modal hallucination control by visual information grounding},
  author={Favero, Alessandro and Zancato, Luca and Trager, Matthew and Choudhary, Siddharth and Perera, Pramuditha and Achille, Alessandro and Swaminathan, Ashwin and Soatto, Stefano},
  booktitle={Proceedings of the IEEE/CVF Conference on Computer Vision and Pattern Recognition},
  pages={14303--14312},
  year={2024}
}

@inproceedings{leng2024mitigating,
  title={Mitigating object hallucinations in large vision-language models through visual contrastive decoding},
  author={Leng, Sicong and Zhang, Hang and Chen, Guanzheng and Li, Xin and Lu, Shijian and Miao, Chunyan and Bing, Lidong},
  booktitle={Proceedings of the IEEE/CVF Conference on Computer Vision and Pattern Recognition},
  pages={13872--13882},
  year={2024}
}

@inproceedings{ghosh2025visual,
  title={Visual description grounding reduces hallucinations and boosts reasoning in lvlms},
  author={Ghosh, Sreyan and Evuru, Chandra Kiran and Kumar, Sonal and Tyagi, Utkarsh and Nieto, Oriol and Jin, Zeyu and Manocha, Dinesh},
  booktitle={International Conference on Learning Representations},
  volume={2025},
  pages={66510--66547},
  year={2025}
}

@article{li2025self,
  title={Self-rewarding vision-language model via reasoning decomposition},
  author={Li, Zongxia and Yu, Wenhao and Huang, Chengsong and Liang, Zhenwen and Liu, Rui and Liu, Fuxiao and Che, Jingxi and Yu, Dian and Boyd-Graber, Jordan and Mi, Haitao and others},
  journal={arXiv preprint arXiv:2508.19652},
  year={2025}
}
\bibliographystyle{plain}

\end{document}